\def\year{2022}\relax
\documentclass[letterpaper]{article} % DO NOT CHANGE THIS
\usepackage{aaai22}  % DO NOT CHANGE THIS
\usepackage[utf8]{inputenc}
\usepackage{amsmath}
\usepackage{amsfonts}
\DeclareMathOperator*{\argmax}{arg\,max}

\usepackage{booktabs}
\usepackage[normalem]{ulem}
\useunder{\uline}{\ul}{}
\usepackage{multirow}
\usepackage{cleveref}
\usepackage{times}  % DO NOT CHANGE THIS
\usepackage{helvet}  % DO NOT CHANGE THIS
\usepackage{courier}  % DO NOT CHANGE THIS
\usepackage[hyphens]{url}  % DO NOT CHANGE THIS
\usepackage{graphicx} % DO NOT CHANGE THIS
\usepackage{natbib}  % DO NOT CHANGE THIS AND DO NOT ADD ANY OPTIONS TO IT
\usepackage{caption} % DO NOT CHANGE THIS AND DO NOT ADD ANY OPTIONS TO IT
\DeclareCaptionStyle{ruled}{labelfont=normalfont,labelsep=colon,strut=off} % DO NOT CHANGE THIS
\usepackage{algorithm}
\usepackage{algorithmic}

\usepackage{newfloat}
\usepackage{listings}
\floatstyle{ruled}
\newfloat{listing}{tb}{lst}{}
\floatname{listing}{Listing}
\title{ASM2TV: An Adaptive Semi-Supervised Multi-Task Multi-View Learning Framework for Human Activity Recognition}
\author{
    Zekai Chen\textsuperscript{\rm 1}, 
    Xiao Zhang\textsuperscript{\rm 2}\thanks{corresponding author},
    Xiuzhen Cheng\textsuperscript{\rm 2}
}
\affiliations{
    \textsuperscript{\rm 1}George Washington University\\
    zech\_chan@gwu.edu
    
    \textsuperscript{\rm 2}School of Computer Science and Technology, Shandong University\\
    \{xiaozhang, xzcheng\}@sdu.edu.cn
}

\usepackage{bibentry}
\begin{document}

\maketitle

\begin{abstract}
Many real-world scenarios, such as human activity recognition (HAR) in IoT, can be formalized as a multi-task multi-view learning problem. Each specific task consists of multiple shared feature views collected from multiple sources, either homogeneous or heterogeneous. Common among recent approaches is to employ a typical hard/soft sharing strategy at the initial phase separately for each view across tasks to uncover common knowledge, underlying the assumption that all views are conditionally independent. On the one hand, multiple views across tasks possibly relate to each other under practical situations. On the other hand, supervised methods might be insufficient when labeled data is scarce. To tackle these challenges, we introduce a novel framework {\it{ASM2TV}} for semi-supervised multi-task multi-view learning. We present a new perspective named gating control policy, a learnable task-view-interacted sharing policy that adaptively selects the most desirable candidate shared block for any view across any task, which uncovers more fine-grained task-view-interacted relatedness and improves inference efficiency. Significantly, our proposed gathering consistency adaption procedure takes full advantage of large amounts of unlabeled fragmented time-series, making it a general framework that accommodates a wide range of applications. Experiments on two diverse real-world HAR benchmark datasets collected from various subjects and sources demonstrate our framework's superiority over other state-of-the-arts. The detailed codes are available at https://github.com/zachstarkk/ASM2TV.
\end{abstract}

\section{Introduction}
\label{introduction}
In IoT mobile sensing world, with the rapidly growing volume of  Internet-connected sensory devices, the IoT generates massive data  characterized by its velocity in terms of spatial and temporal dependency  \cite{Mahdavinejad2018}. 
The sensing data collected from heterogeneous devices are multiple modalities, and multi-task multi-view learning (M2TVL) provides a useful paradigm. 
For instance, in human activity recognition (HAR), sensor monitoring data from various on-body positions can be multi-view. 
% For traffic control in the smart city, traffic data can be collected from cars, road cameras, and counter sensors on roads, where each data source represents a  separate view. 
M2TVL aims to improve 
% efficiency and 
accuracy by learning multiple objectives of tasks with multiple shared view features collected from diverse sources simultaneously \cite{Caruana1998,He2011}. 
% with various views or modalities and varying data quality 
Compared with single-task multi-view learning, the M2TVL paradigm can further improve the training efficiency and reduce inference cost while promoting the generalization effect  \cite{Baxter1997,Ruder2017,Wu2018,Sun2020} by learning shared representations across related tasks and views.  

% Multi-task multi-view learning (M2TVL) aims to improve efficiency and prediction accuracy by learning multiple objectives with multiple shared view features collected from diverse sources simultaneously \cite{Caruana1998,He2011}.
When addressing M2TVL in mobile sensing problems, 
we encounter several challenges. 
(1) Since both views and tasks can be either heterogeneous or homogeneous, one instinctive question is {\it{which views should share across which tasks under what circumstances}} to avoid harmful interference or negative transfer \cite{Kang2011,Standley2020} and optimally reinforce the positive learning. Most prior works \cite{Zhang2012,Jin2013,Jin2014,Lu2017} utilized soft-sharing constraints and considered each view separately by dividing the M2TVL problem into multiple multi-task learning problems under different views while ignoring the more fine-grained {\it{task-view-interacted relatedness}}. 
(2) Also, {\it{as the total number of views and tasks grows, the computation cost proliferates, making the soft-sharing scheme more computationally limited}}. (3) Moreover, {\it{obtaining labeled data in the real mobile sensing world is costly, while the unlabeled data is prevalent and easily accessible}} \cite{Mahdavinejad2018}. Thus, a semi-supervised learning approach is strongly preferred to utilize these unlabeled data adequately. 
% Previous research \cite{} attempted from the perspective of employing a multi-view regularization that enforced all models built on top of each view to agreeing with each other as much as possible on unlabeled data. Nevertheless, we argue this might be inappropriate since different views could be heterogeneous. 
\begin{figure*}[htb]
    \centering
    \includegraphics[width=.7\linewidth]{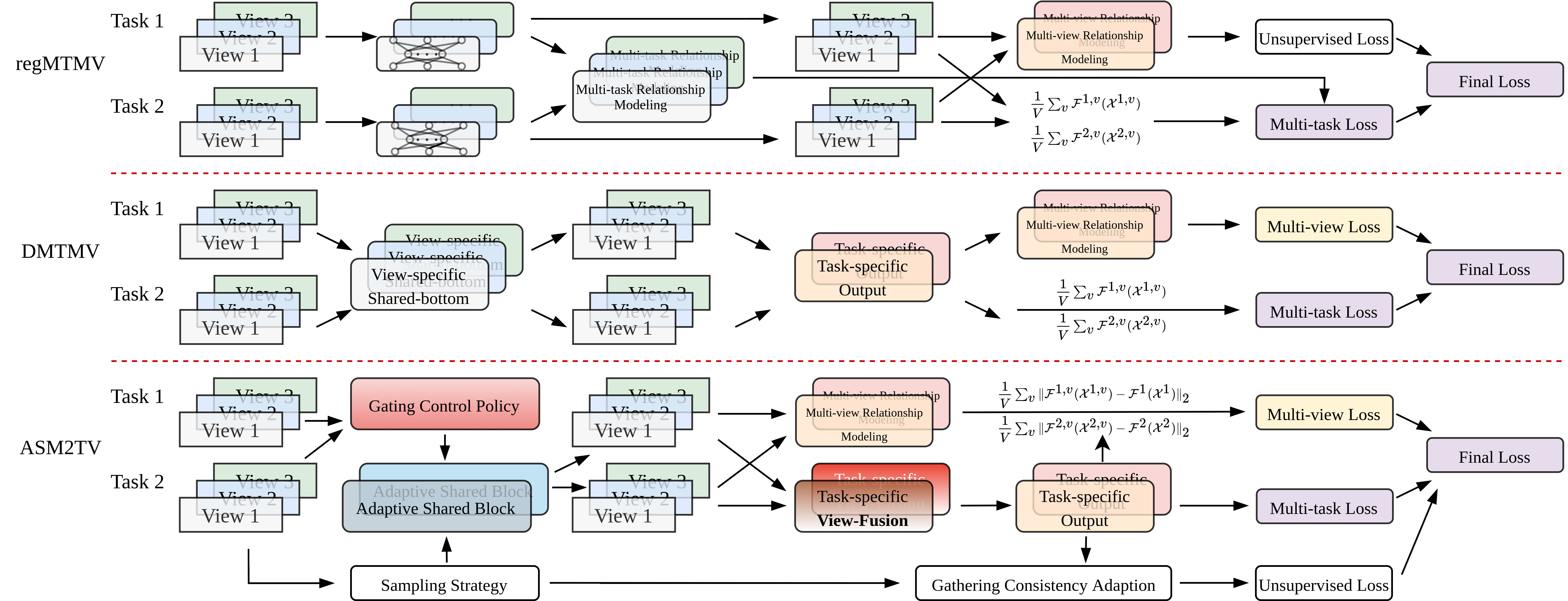}
    \caption{An overview of our framework by comparing with other approaches. For simplicity, we consider an M2TVL scenario with two tasks consisted of three views each. Early M2TV methods use the soft-sharing (e.g., regMTMV) via controlling the similarity between functions built on top of each view across each task. Recent methods such as DMTMV applies the hard-sharing mechanism to share initial bottom layers across tasks under every single view. However, these typical M2TV methods focus modeling relationship on either tasks or views separately while ignoring the task-view-specific relatedness. In our proposed ASM2TV, we design an adaptive gating control policy that learns to determine which views across which tasks should share information to maximize the positive transfer gain combined with a new multi-view regularization based on the view-fusion block. Also, we devise a novel gathering consistency adaption algorithm for semi-supervised learning to take full advantage of unlabeled fragmented time series.}
    \label{fig:main_fig}
    \vspace{-0.4cm}
\end{figure*}

Along this line, we propose a novel semi-supervised multi-task multi-view learning framework {\it{ASM2TV}} to address the above problems (see Figure \ref{fig:main_fig}). Considering computation efficiency, we inherit the spirit of the {\it{hard-parameter sharing}}. Unlike previous M2TVL hard-sharing schemes, we pre-define a set of commonly shared bottoms composed of hidden layers for tasks across all views instead of designing specific shared layers for each view. The main idea is to learn an \emph{adaptive gating control policy} that determines which shared block should be open-gated and others are closed for any view across any task. In other words, the model can adaptively learn what views across which tasks should share common knowledge via the same block to gain the maximum positive transfer benefits. We also design a regularization term that encourages each view-specific function to agree with the view-fusion function as much as possible. Considering most sensor devices generate data continuously and intermittently, we devise a gathering consistency adaption algorithm for semi-supervised learning on fragmented time series. Based on a particular sampling strategy concerning the temporal data characteristics, we further combine it with the consistency training \cite{Bachman2014,Rasmus2015,Laine2017,Xie2020} for unsupervised learning. Additionally, a task-dependent uncertainty modeling strategy is applied to prevent unsupervised loss from being trivial. The main contributions of our work are as follows:
% We use this view-fusion block to inspire individual models built on every single view to discover more generalized information shared among all views; astonishingly, this process feeds back the view-fusion block, generating more accurate predictions as the final outputs.
% Despite many recent works \cite{} have demonstrated relatively astonishing prediction results with fascinating soft-sharing approaches, the model size will also increase proportionally concerning the number of tasks or views. As a result, a soft-sharing scheme might be inappropriate in this scenario, as we target mobile sensing platforms with limited computation resources. 
\begin{itemize}
	\item We propose a novel multi-task multi-view learning framework {\it{ASM2TV}} that automatically learns the sharing schemes across different views and tasks to strengthen the positive learning. 
	\item We design a novel regularization term to enforce models built on every view to agreeing with the view-fusion function, which further benefits the global optimization.
	\item We devise the gathering consistency adaption algorithm combined with a task-dependent uncertainty modeling strategy for semi-supervised learning on fragmented time series. It flexibly takes advantage of a large amount of unlabeled data, making our {\it{ASM2TV}} a general framework for a wide range of applications.
\end{itemize}

\section{Related Work}
\label{related_work}
\paragraph{Multi-Task Learning.}
MTL aims to improve the generalization and performance by weighing the training knowledge among multiple tasks \cite{zhang2018label}. Traditionally, MTL can  model the relationship structures among multiple tasks, for  example, the fully connected structure \cite{evgeniou2004regularized}, tree structure \cite{kim2010tree}, or network structure \cite{chen2012smoothing}. The existing methods of MTL have often been partitioned into two groups with a familiar dichotomy: hard parameter sharing vs. soft parameter sharing. Hard parameter sharing is the practice of sharing model weights between multiple tasks, so that each weight is trained to jointly minimize multiple loss functions \cite{huang2015,kokkinos2017,ranjan2019,bilen2016}. Under soft parameter sharing, different tasks have individual task-specific models with separate weights, but the distance between the model parameters of different tasks is added to the joint objective function, such as Cross-stitch \cite{Misra2016}, Sluice \cite{ruder2019} and NDDR \cite{gao2019}, consist of a network column for each task, and define a mechanism for feature sharing between columns.
% Though there is no explicit parameter sharing, there is an incentive for the task-specific models to have similar parameters. 
% However, the mentioned multi-task learning models can not deal with 

% One focus of MTL is to model the relationships among multiple tasks. 

% When the correlations among tasks are weak, the above method may lead to negative transfer, 
% and the model can be constrained by introducing task  clustering\cite{evgeniou2004regularized}. 

% \cite{caruana1997multitask} The  earliest MTL method \cite{caruana1997multitask} shared some  parameters in all tasks (general bottom layer) and used its own  unique parameters in specific task layer (top layer). 

% MTL in the  traditional method mainly focuses on two points: making the model  sparse among tasks through norm regularization, and modeling the  relationship between multiple tasks.  \cite{argyriou2007multi}\cite{lounici2009taking}\cite{jalali2010dirty}\cite{liu2017distributed} forces the model to consider only some features, provided that different tasks must be related. 

% When the correlation between tasks is weak, the above method may lead to negative transfer (that is, negative effect), and the model can be constrained by introducing task  clustering\cite{evgeniou2004regularized}.  \cite{thrun1996discovering}\cite{lawrence2004learning}\cite{bakker2003task}\cite{kang2011learning}are other methods used to learn task relationship. 
% These methods can learn some shared features or shared structures between different tasks, but they cannot share information between different views. 

\paragraph{Multi-Task Multi-View Learning.}
The research on multi-task multi-view learning has attracted wide attention in recent years. GraM2  \cite{He2011} proposed a graph-based framework, in which an effective algorithm was proposed to optimize the framework. However, GraM2 can only deal with the non-negative feature values. The regMVMT \cite{Zhang2012} algorithm was proposed based on the idea of co-regularization, assuming that different view of prediction models should be consistent. Along this line, a more generalized algorithm CSL-MTMV \cite{Jin2013} was proposed based on the assumption that multiple related tasks with the same view should be shared in the low-dimensional common subspace. MAMUDA \cite{Jin2014} was proposed for heterogeneous tasks, in which the shared structure and task-specific structures can be combined into a unified formulation. DMTMV \cite{Wu2018} was a 
% unified deep 
multi-task multi-view learning framework including shared feature network, specific feature network, and task network. Nevertheless, these methods assume independence among views and tasks, which is improper in many real-life scenarios. Therefore, we aim to propose a new framework to address these problems. 

\paragraph{Human Activity Recognition.} 
% 	Multiple sensors based heterogeneous human activity recognition (HAR) has drawn many researchers' interests in the past few years. 
	The existing approaches for sensor based HAR usually adopted kinds of deep learning techniques, for instance,  LSTM learners  \cite{mv-rnn-1} or combinations of recurrent models and CNN  networks \cite{deepSense}\cite{attnsense} \cite{har-iprnn}. 
	However, none of the previous methods addressed HAR from a semi-supervised multi-task multi-view learning perspective, which is the focus of our work. 

\section{ASM2TV: Adaptive Semi-Supervised Multi-Task Multi-View Learning}
\label{method}
Given a set of $T$ tasks with each task containing $V$ different views, we define $\mathcal{X}^{t, v}$ as the original inputs from $v$-th view in $t$-th task. We aim to seek an adaptive sharing scheme that best describes the internal task-view-interacted relatedness \cite{Jin2014,Lu2017} instead of investigating task-task or view-view association. From the perspective of hard-sharing, we seek to know which views should be shared across which tasks to optimally augment the positive learning while avoiding the negative transfer \cite{Kang2011,Standley2020}. The computation efficiency for scalable M2TVL is also considered as we mainly target IoT mobile computing scenarios. 

% Though general RNN (recurrent neural network) blocks own superiority on exhibiting dynamic temporal sequences, it is yet to be demonstrated effective in modeling short, fragmented time series since the internal state can barely obtain sufficient information to make accurate predictions. This paper considers using the basic convolution blocks or simple feedforward linear blocks as the candidate shared blocks across fragmented time series from all views and tasks. 

\subsection{Task-View-Interacted Gating Control Policy}
\begin{figure}
    \centering
    \includegraphics[width=.6\linewidth]{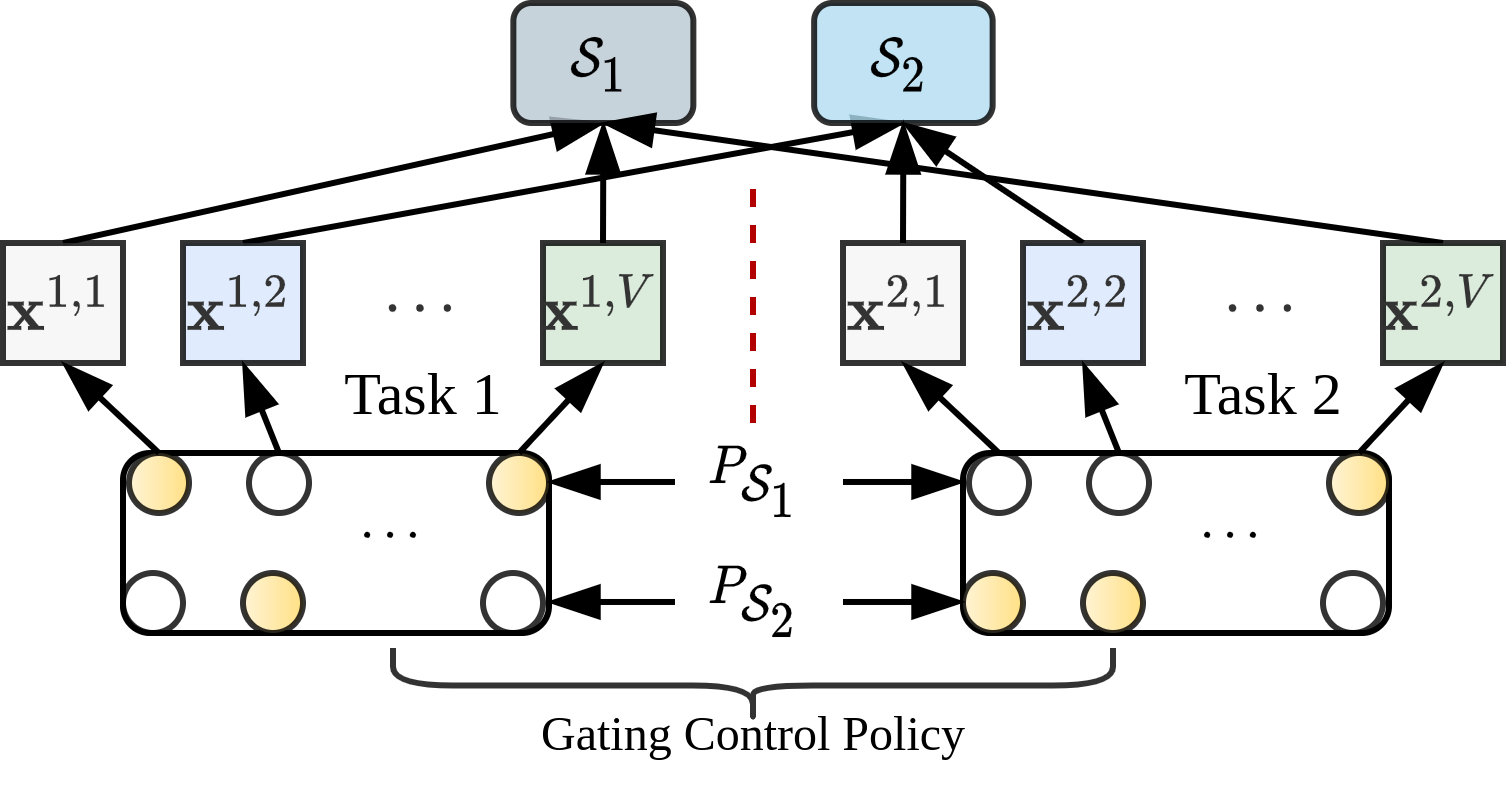}
    \caption{Suppose we have two tasks consisted of $V$ views each and two shared blocks $\mathcal{S}_{1}$ and $\mathcal{S}_{2}$. Our gating control policy's main idea is to use the Gumbel-Softmax Sampling strategy to sample a random categorical vector for determining which views across which tasks to share through the same shared block. For $\mathbf{x}^{1, v}$, since $\mathcal{P}_{\mathcal{S}_{1}} > \mathcal{P}_{\mathcal{S}_{2}}$ (yellow circles), the first shared block will be chosen to execute.}
    \label{fig:gating}
    \vspace{-0.4cm}
\end{figure}

Following the spirit of {\it{hard-parameter sharing}}, we pre-define a set of initially shared blocks (or shared bottoms) for any input from each view and task to potentially execute or share across (see Figure \ref{fig:gating}). We use a random categorical variable $\mathbf{z}^{t, v}$  that determinates whether blocks are open-gated to any inputs $\mathcal{X}^{t, v}$ from the $v$-th view of the $t$-th task. It can also be viewed as a soft-clustering process since similar representations attempt to share across the same block (see Figure \ref{fig:correlation}). As long as the learned pattern produce \textbf{positive} feedback, different views across all tasks would be encouraged to share with each other, making it no longer limited to knowledge sharing under the same view. Inspired by \cite{Maddison2014,Jang2017}, we adopt the Gumbel-Softmax Sampling technique to optimize our gating control policy jointly with the model parameters $\theta$ through standard backpropagation. Instead of constructing a policy network to form a specific policy for each input mini-batch, we employ a universal learnable policy to make structural decisions to evade the explosion of parameter complexity restricted by the computation resources. 

\paragraph{Gumbel-Softmax Sampling Trick.} The sampling process of discrete data from a categorical distribution is originally non-differentiable, where typical backpropagation in deep neural networks cannot be conducted. \cite{Maddison2014,Jang2017} proposed a differentiable substitution of discrete random variables in stochastic computations by introducing Gumbel-Softmax distribution, a continuous distribution over the simplex that can approximate samples from a categorical distribution. In our gating control policy with a total number of $N$ candidate shared blocks, we let $\mathbf{z}^{t, v}$ be the gating control variable for inputs $\mathcal{X}^{t, v}$ with open-gate probabilities for each block as $\pi^{t, v}_{1}, \cdots, \pi^{t, v}_{N}$, where $\mathcal{X}^{t, v}$ are inputs from the $v$-th view of the $t$-th task and $\pi^{t, v}_{i}, \forall i\in \{1, \cdots, N\}$ represents the probability that the $i$-th shared block would be open to $\mathcal{X}^{t, v}$. Similarly, by Gumbel-Max trick, we can sample any block's open-or-not gating strategy $z^{t, v}$ for inputs $\mathcal{X}^{t, v}$ with:
\begin{equation}
    z^{t, v} = \argmax_{i}(\log \pi^{t, v}_{i} + g^{t, v}_{i})
\end{equation}
where $g_{i}, \cdots, g_{N}$ are i.i.d samples drawn from a standard Gumbel distribution which can be easily sampled using inverse transform sampling by drawing $u \sim$ Uniform$(0, 1)$ and computing $g=-\log(-\log u)$. We further substitute this $\argmax$ operation, since it is not differentiable, with a $\textsc{Softmax}$ reparameterization trick, also known as Gumbel-Softmax trick, as:
\begin{equation}
    z^{t, v}_{i} = \frac{\exp((\log \pi^{t, v}_{i} + g^{t, v}_{i})/\tau)}{\sum_{j=1}^{N}\exp((\log \pi^{t, v}_{j} + g^{t, v}_{j})/\tau)}
\end{equation}
where $i\in \{1, \cdots, N\}$ and $\tau$ is the temperature parameter to control Gumbel-Softmax distribution's smoothness, as the temperature $\tau$ approaches 0, the Gumbel-Softmax distribution becomes identical to the one-hot categorical distribution. As the randomness of $g$ is independent of $\pi$, we can now directly optimize our gating control policy using standard gradient descent algorithms. 

% let $\mathbf{z}$ be a categorical variable with class probabilities $\pi_{1}, \pi_{2}, \cdots, \pi_{N}$. The Gumbel-Max trick \cite{} provides a simple and effective way to draw samples $z$ from the following distribution with class probability $\pi$:
% \begin{equation}
%     z = \argmax_{i}(\log \pi_{i} + g_{i})
% \end{equation}
% where $g_{i}, \cdots, g_{N}$ are i.i.d samples drawn from a standard Gumbel distribution which can be easily sampled using inverse transform sampling by drawing $u \sim$ Uniform$(0, 1)$ and computing $g=-\log(-\log u)$. In our gating control policy with a total number of $N$ candidate shared blocks, 

\paragraph{View-Fusion for Co-Regularization.}
Existing approaches attempt to tackle the M2TV problem as multiple multi-task learning problems from different views, underlying the assumption that all views are conditionally independent \cite{Zhang2012,Jin2013,Jin2014,Wu2018}. If we let $\mathcal{F}^{v}$ represent the view function built on view $v$, a typical final model obtained on one task $t$ is the average of prediction results from all views (also see Figure \ref{fig:main_fig}) as
$\mathcal{F}(\mathcal{X}) = \frac{1}{V}\sum_{v=1}^{V}\mathcal{F}^{v}(\mathcal{X}^{v})$, 
where $V$ is the total number of views, $\mathcal{X}$ is the original multi-task multi-view inputs, and $\mathcal{X}^{v}$ is the set of features containing data from all tasks under view $v$. Unlike the above, we encourage the function built on each view to agreeing with the view-fusion module compiled on merged views. In our work, we use a specific feedforward linear layer as the view-fusion block to merge multiple views by taking all views concatenated. Thus, we design a regularization term as:
\begin{equation}
    \mathcal{L}_{f} = \sum_{t}\sum_{v}\frac{1}{V}\Vert \mathcal{F}^{t, v}(\mathcal{X}^{t, v}) - \mathcal{F}^{t}(\mathcal{X}^{t})\Vert_{2} 
\end{equation}
where $\mathcal{F}^{t, v}$ is the task-view-specific model function for inputs $\mathcal{X}^{t, v}$ from the $v$-th view of $t$-th task, $\mathcal{F}^{t}$ is the view-fusion function for inputs $\mathcal{X}^{t}$ from task $t$ with all views merged. 

\subsection{Semi-Supervised Learning for Time Series}
Considering temporal data characteristics from mobile sensors, we argue that the time series should always reflect an individual's coherent and unified physical status within a relatively short interval. Thus, we expect adjacent sub time series within the same time interval (aka. from internal time interval) to obtain similar representations. In contrast, time series distant from each other (aka. from external time interval) may obtain distinctive representations.
\begin{algorithm}
\algsetup{linenosize=\tiny}
  \scriptsize
\caption{\small{Semi Supervised Multi-Task Multi-View Learning for Fragmented Time Series with Gathering Consistency Adaption}}
% \In{Time series data \mathcal{x}_{s}}
\textbf{Input}: Labeled multi-task multi-view time series data $\mathcal{X}_{s}$, unlabeled data $\mathcal{X}_{u}$, number of tasks $T$, model $\mathcal{F}$ with parameters $\theta$, adaption steps $K$, supervised loss $\mathcal{L}_{s}$, unsupervised loss $\mathcal{L}_{u}$, unsupervised loss coefficient $\lambda$, uncertainty weights parameter $\alpha_{t}, \beta_{t}$ for any task $t$, learning rate $\eta$\\
\textbf{Output}: Model parameters $\theta^{'}$
\begin{algorithmic}[1] %[1] enables line numbers
 \STATE Let $t\in \{1, \cdots, T\}$ represents the specific task code, the current 
 \STATE Initialization $\theta$ and $\{\beta_{t}\}$
 \WHILE{Training}
  \STATE Let $\mathbf{x}_{s}=\{\mathbf{x}_{s}^{t}\}$ be a mini-batch of $\mathcal{X}_{s}$
  \STATE For all tasks, randomly select subseries of $\mathcal{X}_{u}$ from the internal time slot as original unlabeled inputs $\mathbf{x}_{u} = \{\mathbf{x}_{u}^{t}\}_{t\in \{1, \cdots, T\}}$ for reference
  \STATE For all tasks, select other subseries of $\mathcal{X}_{u}$ uniformly at random from the internal time slot for $K$ times as $\widehat{\mathbf{x}}_{u} = \{\widehat{\mathbf{x}}_{u, 1}, \cdots, \widehat{\mathbf{x}}_{u, K}\}$, in which for $\forall{k} \in \{1, \cdots, K\}, \widehat{\mathbf{x}}_{u, k}=\{\widehat{\mathbf{x}}_{u, k}^{t}\}_{t\in \{1, \cdots, T\}}$
  \STATE For all tasks, select two subseries from the external time slot, one before the internal and one after, as $\tilde{\mathbf{x}}_{u} = \{\tilde{\mathbf{x}}_{u, 1}, \tilde{\mathbf{x}}_{u, 2}\}$, where $\tilde{\mathbf{x}}_{u, i} = \{\tilde{\mathbf{x}}^{t}_{u, 1}\}_{t\in \{1, \cdots, T\}}, \forall i \in [1, 2]$
  \STATE $[\mathbf{y}_{s}, \widehat{\mathbf{y}}_{u}, \tilde{\mathbf{y}}_{u}] \leftarrow \mathcal{F}(\mathbf{x}_{s}, \widehat{\mathbf{x}}_{u}, \tilde{\mathbf{x}}_{u}; \theta)$
  \STATE $\mathbf{y}_{u} \leftarrow \mathcal{F}(\mathbf{x}_{u}; \tilde{\theta})$, $\tilde{\theta}$ is a hard copy of $\theta$
  \STATE Update $\mathcal{L}_{u}$ based on Eq. \ref{eq:objective_function}
  \STATE $\mathcal{J}(\theta) \leftarrow \mathcal{L}_{s}(\mathcal{X}_{s}; \theta)+\lambda \mathcal{L}_{u}(\mathcal{X}_{u}; \theta, \tilde{\theta})$
  \STATE $\theta^{'} \leftarrow \theta - \eta \nabla_{\theta}\mathcal{J}(\theta) $
 \ENDWHILE
 \end{algorithmic}
 \label{alg:semi}
\end{algorithm}

\begin{figure}
    \centering
    \includegraphics[width=.9\linewidth]{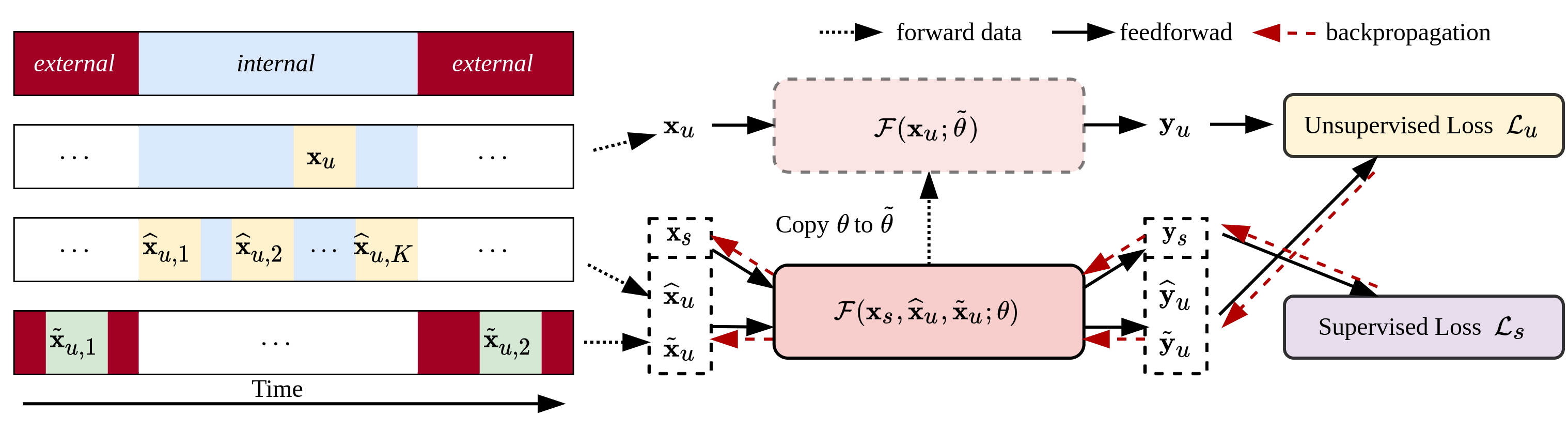}
    \caption{The semi-supervised learning framework shown in Algorithm \ref{alg:semi}}
    \label{fig:semi}
    \vspace{-0.4cm}
\end{figure}

 \paragraph{Gathering Consistency Adaption.} To accommodate this principle to mobile sensing time series, we proposed this gathering consistency adaption algorithm, as described in Algorithm \ref{alg:semi} and Figure \ref{fig:semi}, for semi-supervised learning. We utilize an original sampling strategy combined with the consistency training \cite{Bachman2014,Laine2017,Franceschi2019,Xie2020} on numerous unlabeled time series to constrain model predictions to be invariant to similar inputs with inevitable noise while sensitive to essential differentiation. Let $\mathcal{X}_{s}$ and $\mathcal{X}_{u}$ represents the overall labeled data and unlabeled data, respectively. We split any given long time series from $\mathcal{X}_{u}$ into multiple fragmented time series for convenience. Intuitively, for every training step, the chosen fragmented sequence is viewed as an internal time slot while other sequences are external time slots. We first select a random subseries $\mathbf{x}_{u}$ within the internal time slot as the original reference. We then uniformly select multiple random subseries $\{\widehat{\mathbf{x}}_{u, 1}, \cdots, \widehat{\mathbf{x}}_{u, K}\}$ as a set of similar but naturally noisy samples compared to $\mathbf{x}_{u}$ within the same internal time interval for consistency regularization. Finally, we also randomly select two other subseries $\{\tilde{\mathbf{x}}_{u, 1}, \tilde{\mathbf{x}}_{u, 2}\}$ from the external slots, respectively before and after the internal slot in time for inconsistency differentiation. Therefore, we name this approach as gathering consistency adaption since it is a cumulative loss adaption process, including consistency and inconsistency training. In this case, we apply $\textsc{KL-Divergence}$ as the divergence metric between any pair of probability distributions $\mathcal{D}(P_{\theta}(y\vert x_{u})\Vert P_{\theta}(y\vert \widehat{x}_{u}, \tilde{x}_{u}))$. Figure \ref{fig:semi} is a visualization of the semi-supervised learning architecture. 

We calculate a cumulative loss through multiple adaption steps, yet the $\textsc{KL-Divergence}$ is still too trivial to use directly. Also, the loss between different representations can effortlessly dominate the loss between similar ones. To tackle this challenge, we extend the task-dependent uncertainty modeling approach inspired by \cite{Kendall2018}.

\begin{table*}[htb]
\centering
\vspace{-.5cm}
\caption{Prediction results on {\bf{RealWorld-HAR}} (8 tasks 7 views) with best model performance in bold and second-best results with underlines. {\it{ASM2TV}} generally achieves the best performance and prediction results on all metrics across all exhibited tasks. $\Delta_{\mathcal{T}}$ represents the improvement percentage compared to the second-best approach for all tasks. Our framework achieves a significant improvement compared to other state-of-the-arts.}
\resizebox{.6\linewidth}{!}{%
\begin{tabular}{@{}lccccccccc@{}}
\toprule
\multirow{2}{*}{Proband} & \multirow{2}{*}{Metrics} & \multicolumn{3}{c}{Supervised Models} & \multicolumn{4}{c}{Semi-Supervised Models} & \multicolumn{1}{c}{\multirow{2}{*}{\begin{tabular}[c]{@{}l@{}}$\Delta \uparrow$\\ ($\%$)\end{tabular}}} \\ \cmidrule(lr){3-5} \cmidrule(lr){6-9} 
 &  & \multicolumn{1}{l}{DeepSense} & \multicolumn{1}{l}{IteMM} & \multicolumn{1}{l}{DMTMV} & \multicolumn{1}{l}{reg-MTMV} & \multicolumn{1}{l}{CSL-MTMV} & \multicolumn{1}{l}{MFMs} & \multicolumn{1}{l}{Ours} & \multicolumn{1}{l}{} \\ \midrule
\multirow{4}{*}{$\mathcal{T}_{1}$} & Acc & 0.5667 & 0.2465 & {\ul 0.6937} & 0.5312 & 0.5337 & 0.6932 & \textbf{0.7915} & 14.10\% \\
 & M-F1 & 0.4949 & 0.2639 & 0.6762 & 0.5317 & 0.5121 & {\ul 0.6816} & \textbf{0.7612} & 11.68\% \\
 & W-F1 & 0.5361 & 0.2571 & 0.6678 & 0.5211 & 0.5313 & {\ul 0.6994} & \textbf{0.7493} & 7.13\% \\
 & Avg & 0.5326 & 0.2558 & 0.6792 & 0.5280 & 0.5257 & {\ul 0.6914} & \textbf{0.7673} & 10.98\% \\
\multirow{4}{*}{$\mathcal{T}_{2}$} & Acc & 0.7131 & 0.3841 & 0.8361 & 0.7649 & 0.7745 & {\ul 0.8237} & \textbf{0.8659} & 5.12\% \\
 & M-F1 & 0.6353 & 0.4532 & 0.7031 & 0.7505 & 0.7614 & {\ul 0.8127} & \textbf{0.8662} & 6.58\% \\
 & W-F1 & 0.7076 & 0.4319 & 0.7851 & 0.7694 & 0.7794 & {\ul 0.8263} & \textbf{0.8416} & 1.85\% \\
 & Avg & 0.6853 & 0.4231 & 0.7748 & 0.7616 & 0.7718 & {\ul 0.8209} & \textbf{0.8579} & 4.51\% \\
\multirow{4}{*}{$\mathcal{T}_{3}$} & Acc & 0.5829 & 0.2633 & 0.7374 & 0.6962 & 0.7362 & {\ul 0.7945} & \textbf{0.8474} & 6.66\% \\
 & M-F1 & 0.4659 & 0.2798 & 0.7381 & 0.6986 & 0.7276 & {\ul 0.7545} & \textbf{0.8073} & 7.00\% \\
 & W-F1 & 0.5412 & 0.2726 & 0.7343 & 0.6832 & 0.7306 & {\ul 0.7497} & \textbf{0.7875} & 5.04\% \\
 & Avg & 0.5300 & 0.2719 & 0.7366 & 0.6927 & 0.7315 & {\ul 0.7662} & \textbf{0.8141} & 6.24\% \\
\multirow{4}{*}{$\mathcal{T}_{4}$} & Acc & 0.7046 & 0.2793 & {\ul 0.7222} & 0.5287 & 0.5291 & 0.6891 & \textbf{0.7911} & 9.54\% \\
 & M-F1 & 0.5631 & 0.3479 & {\ul 0.6557} & 0.5279 & 0.5193 & 0.6429 & \textbf{0.7152} & 9.07\% \\
 & W-F1 & 0.6743 & 0.3034 & {\ul 0.6961} & 0.5295 & 0.5279 & 0.6632 & \textbf{0.7583} & 8.94\% \\
 & Avg & 0.6473 & 0.3102 & {\ul 0.6913} & 0.5287 & 0.5254 & 0.6651 & \textbf{0.7549} & 9.19\% \\
\multirow{4}{*}{$\mathcal{T}_{5}$} & Acc & 0.5020 & 0.2991 & {\ul 0.7461} & 0.5068 & 0.518 & 0.657 & \textbf{0.8163} & 9.41\% \\
 & M-F1 & 0.4867 & 0.3741 & {\ul 0.7458} & 0.4735 & 0.4742 & 0.6411 & \textbf{0.7815} & 4.79\% \\
 & W-F1 & 0.4714 & 0.3269 & {\ul 0.7388} & 0.4674 & 0.478 & 0.6429 & \textbf{0.7665} & 3.75\% \\
 & Avg & 0.4867 & 0.3334 & {\ul 0.7436} & 0.4826 & 0.4901 & 0.6470 & \textbf{0.7881} & 5.99\% \\
\multirow{4}{*}{$\mathcal{T}_{6}$} & Acc & 0.6798 & 0.3689 & \textbf{0.8329} & 0.726 & 0.7264 & 0.8066 & {\ul 0.8161} & -2.02\% \\
 & M-F1 & 0.6771 & 0.451 & \textbf{0.793} & 0.6781 & 0.6784 & 0.7461 & {\ul 0.7581} & -4.40\% \\
 & W-F1 & 0.6743 & 0.4287 & \textbf{0.7873} & 0.6713 & 0.6819 & 0.755 & {\ul 0.7619} & -3.23\% \\
 & Avg & 0.6771 & 0.4162 & \textbf{0.8044} & 0.6918 & 0.6956 & 0.7692 & {\ul 0.7787} & -3.19\% \\
\multirow{4}{*}{$\mathcal{T}_{7}$} & Acc & 0.4883 & 0.302 & {\ul 0.7481} & 0.6597 & 0.6619 & 0.7263 & \textbf{0.7921} & 5.88\% \\
 & M-F1 & 0.4675 & 0.3672 & {\ul 0.6941} & 0.6263 & 0.626 & 0.689 & \textbf{0.7181} & 3.46\% \\
 & W-F1 & 0.4466 & 0.3175 & {\ul 0.703} & 0.6196 & 0.6349 & 0.6926 & \textbf{0.7374} & 4.89\% \\
 & Avg & 0.4675 & 0.3289 & {\ul 0.7151} & 0.6352 & 0.6409 & 0.7026 & \textbf{0.7492} & 4.77\% \\
\multirow{4}{*}{$\mathcal{T}_{8}$} & Acc & 0.6170 & 0.2916 & {\ul 0.7234} & 0.675 & 0.6752 & 0.7587 & \textbf{0.7707} & 6.54\% \\
 & M-F1 & 0.6019 & 0.3584 & \textbf{0.7043} & 0.6407 & 0.6413 & 0.6433 & {\ul 0.6843} & -2.84\% \\
 & W-F1 & 0.5867 & 0.2994 & {\ul 0.712} & 0.6412 & 0.6697 & 0.6607 & \textbf{0.7309} & 2.65\% \\
 & Avg & 0.6019 & 0.3165 & {\ul 0.7132} & 0.6523 & 0.6621 & 0.6876 & \textbf{0.7286} & 2.16\% \\ \bottomrule
\end{tabular}%
}

\label{tab:main_table}
% \vspace{-0.3cm}
\end{table*}

\begin{table}[htb]
\centering
\caption{Our ASM2TV framework averaged prediction results over all metrics across all tasks compared with other methods on \textbf{GLEAM} dataset.}
\resizebox{.8\linewidth}{!}{%
\begin{tabular}{@{}lccccc@{}}
\toprule
 & Model & \begin{tabular}[c]{@{}c@{}}$\Delta_{Acc} \uparrow$\\ (\%)\end{tabular} & \begin{tabular}[c]{@{}c@{}}$\Delta_{M-F1} \uparrow$\\ (\%)\end{tabular} & \begin{tabular}[c]{@{}c@{}}$\Delta_{W-F1} \uparrow$\\ (\%)\end{tabular} & \begin{tabular}[c]{@{}c@{}}$\Delta_{\#Params} \downarrow$ \\ (\%)\end{tabular} \\ \midrule
\multirow{3}{*}{Supervised} & DeepSense & 0.7216 & 0.7016 & 0.7116 & 10M \\
 & IteMM & 0.3969 & 0.3859 & 0.3985 & - \\
 & DMTMV & 0.8443 & 0.8209 & 0.8397 & {\ul -20\%} \\ \midrule
\multirow{4}{*}{Semi-Supervised} & regMVMT & 0.7432 & 0.7297 & 0.7401 & -0\% \\
 & CSL-MTMV & 0.7721 & 0.7507 & 0.7685 & -0\% \\
 & MFMs & {\ul 0.8515} & {\ul 0.8209} & {\ul 0.8468} & -0\% \\
 & ASM2TV (ours) & \textbf{0.9742} & \textbf{0.9401} & \textbf{0.9464} & \textbf{-47\%} \\ \bottomrule
\end{tabular}%
}
\vspace{-0.4cm}
\label{tab:table4}
\end{table}

\paragraph{Task-Dependent Uncertainty Modeling.} \cite{Kendall2018} derived a multi-task loss function based on maximizing the Gaussian likelihood with homoscedastic uncertainty for weighting losses in multi-task learning. Specifically, for a given multi-task model $\mathcal{F}$ with parameters $\theta$, it leads to an overall minimization loss objective function as:
\begin{equation}
    \mathcal{L}(\theta, \sigma_{1}, \sigma_{2}) \propto \frac{1}{2\sigma_{1}^{2}}\mathcal{L}_{1}(\theta) + \frac{1}{2\sigma_{2}^{2}}\mathcal{L}_{2}(\theta) + \log \sigma_{1}\sigma_{2}
\end{equation}
where $\mathcal{L}_{1}$ and $\mathcal{L}_{2}$ are the loss functions for task $1$ and $2$, underlying the assumption that the prediction probabilities of any given model denote to a Gaussian distribution with mean given by the model output and an observation noise scalar $\sigma$:
\begin{equation}
    P_{\theta}(y\vert \mathcal{F}(x)) = \mathcal{N}(\mathcal{F}(x), \sigma^{2})
\end{equation}
Intuitively, if one particular task's noise effect enlarges, the overall loss for that task would be balanced adaptively to prevent other tasks training from being dominated. In our consistency adaption procedure, the divergence distance between any pair of probability distributions genuinely denotes the Gaussian distribution following the fact that most of the time series are with Gaussian white noise \cite{Mahdavinejad2018}, especially for the IoT sensing data. As a result, the assumption above matches the facts, which makes the uncertainty modeling approach for multi-task semi-supervised loss suitable in this case. We let $\alpha_{t}$ and $\beta_{t}$ (which are equivalent to $2\log \sigma_{t}$) be the learnable noise parameter for consistency loss and discrimination loss of task $t$, respectively. Thus, the minimization objective function becomes:
% _{y_{1}\sim \mathbf{y}^{t}_{u}, y_{2}\sim{\widehat{\mathbf{y}}^{t}_{u, k}}}
% _{y_{1}\sim \mathbf{y}^{t}_{u}, y_{2}\sim{\widehat{\mathbf{y}}^{t}_{u, i}}}
\begin{tiny}
\begin{multline}
\label{eq:objective_function}
    \mathcal{L}_{u}(\theta, \alpha, \beta) = \sum_{t}^{T}\sum_{k}^{K} e^{-\alpha_{t}} \mathbb{E} [\log P_{\tilde{\theta}}(y_{1}\vert \mathbf{x}^{t}_{u}) - 
    \log P_{\theta}(\widehat{y}_{2}\vert \widehat{\mathbf{x}}^{t}_{u, k})] + \alpha_{t}\\
    + \sum_{t}^{T}\sum_{i=1}^{2}-e^{-\beta_{t}} \mathbb{E} [\log P_{\tilde{\theta}}(y_{1}\vert \mathbf{x}^{t}_{u}) - \log P_{\theta}(\tilde{y}_{2}\vert \tilde{\mathbf{x}}^{t}_{u, i})] + 
    \beta_{t}
\end{multline}

\end{tiny}

As the consistency loss is more trivial than the differentiation loss, the learnable variance scalar may also become smaller for the consistency piece. In contrast, the weights for the consistency part increase, which lowers the risk of being dominated by other significant task losses. 

\section{Experiments}
In this section, we conduct experiments on two real-world human activity recognition datasets to show that our model outperforms many strong M2TV baselines, meanwhile maintaining a low-memory footprint for computation efficiency. 

\subsection{Datasets and Descriptions}
We evaluate our framework using two real-world human activity recognition (HAR) datasets, namely {\bf{RealWorld-HAR}} \cite{Sztyler2016}, this data set covers various mobile sensor level data of the activities (e.g., climbing stairs down and up, jumping, etc.) of fifteen probands (or subjects) each for 10 minutes roughly. We take each proband as a separate {\bf{task}} with different on-body positions (e.g., chest, forearm, etc.) recognized as different {\bf{views}}. {\bf{GLEAM}} \cite{Rahman2015}, this dataset is a head-motion-tracking dataset collected with Google Glass with the labeled data from the head-mounted sensor that can be used to recognize eating and other activities, and ultimately to assist individuals with diabetes. We also take each subject as an individual {\bf{task}}, and each sensor is recognized as a single {\bf{view}}. For both datasets, we consecutively split the original multivariate time series into multiple fragmented time series. Each piece lasts only 5 seconds to meet the requirements of instant response under most mobile computing scenarios. For data split, we uniformly separate the time series in a consecutive manner for each specific activity of each subject, such that we can make sure to predict each category sufficiently. For example, for subject A, we chronologically select the 10\% continuous series as labeled data for training, 40\% as unlabeled data, 10\% data with the label for validation, and the rest labeled 40\% series for testing. Noticeably, the split strategy is applied to each activity of each subject consistently. We then transformed each time series into sliding windows with a length of 5 seconds to meet the requirements of instant response under most mobile computing scenarios. Such that, we can utilize a semi-supervised learning approach to evaluate the effectiveness. 

\subsection{Baselines and Metrics}
We compare our method with the following baselines. First, we consider using {\bf{DeepSense}} \cite{Yao2017}, a deep learning model designed for time-series mobile sensing data, as a single-task baseline where we train each task separately. {\bf{IteM2}} \cite{He2011} is a transductive algorithm, and it can only handle non-negative feature values. When applying the IteM2 algorithm to some of our datasets that have negative feature values, we add a positive constant to guarantee the non-negativity; {\bf{regMVMT}} algorithm \cite{Zhang2012} is an inductive algorithm, which assumes all tasks should be similar to achieve a good performance; {\bf{CSL-MTMV}} is an inductive M2TVL algorithm \cite{Jin2013} that assumes the predictions of different views within a single task are consistent; {\bf{MFMs}} is a multilinear factorization model proposed by \cite{Lu2017} which can learn the task-specific feature map. {\bf{DMTMV}} \cite{Wu2018} is a deep multi-task multi-view method to learn nonlinear feature representations and classifier in a unified framework, which can also learn task relationships in nonlinear models. 

To investigate the performance of models, we adopt accuracy (Acc), macro-F1 score (M-F1), and weighted-F1 (W-F1) score on the test data as the evaluation metrics. The larger value of each metric indicates better performance. 

\subsection{Experiment Settings}
We apply the basic feedforward multilayer perceptron (MLP) as the backbone for each task-specific and view-specific layer. The shared blocks are also composed of primary linear hidden layers. We use the {\sc{Cross Entropy}} as the supervised loss for human activity classification. The unsupervised loss can be viewed as an extension of {\sc{KL-Divergence}} loss. We use Adam as the optimizer for all deep neural network-based models and set the initial learning rate to $3e^{-4}$ with a weight decay of $1e^{-6}$. Specifically, an upsampling strategy is applied to overcome the label imbalance problem. We utilize a dropout strategy for all deep neural networks with a dropout rate of $0.5$ to prevent overfitting. For more setting details, readers may refer to the supplementary materials. 

\begin{table}[htb]
\centering
\resizebox{.8\linewidth}{!}{%
\begin{tabular}{@{}lccccc@{}}
\toprule
 &
  \multicolumn{1}{c}{Model} &
  \begin{tabular}[c]{@{}c@{}}$\Delta_{Acc} \uparrow$\\ (\%)\end{tabular} &
  \begin{tabular}[c]{@{}c@{}}$\Delta_{M-F1} \uparrow$ \\ (\%)\end{tabular} &
  \begin{tabular}[c]{@{}c@{}}$\Delta_{W-F1} \uparrow$ \\ (\%)\end{tabular} &
  \begin{tabular}[c]{@{}c@{}}$\Delta_{\#Params} \downarrow$ \\ (\%)\end{tabular} \\ \midrule
\multirow{3}{*}{Supervised}      & DeepSense     & 0.5988 (-)    & 0.5760 (-)    & 0.5797 (-)    & 10M (-) \\
                                 & IteMM         & -49\%         & -49\%         & -48\%         & -        \\
                                 & DMTMV         & {\ul 25\%}    & 24\%          & 25\%          & {\ul -20\%}   \\ \midrule
\multirow{4}{*}{Semi-Supervised} & regMVMT       & 7\%           & 6\%           & 6\%           & -0\%    \\
                                 & CSL-MTMV      & 8\%           & 11\%          & 10\%          & -0\%    \\
                                 & MFMs          & 24\%          & {\ul 25\%}    & {\ul 26\%}    & -0\%    \\
                                 & {\it{ASM2TV}} (ours) & \textbf{46\%} & \textbf{43\%} & \textbf{43\%} & \textbf{-47\%}   \\ \bottomrule
\end{tabular}%
}
\caption{Overview of comparison on {\bf{RealWorld-HAR}}. We additionally exhibit average prediction results of all tasks and display the overall improvement percentage through all quantitative metrics and model space complexity ($\Delta_{\#Params}$). Generally, our {\it{ASM2TV}} can achieve over $40\%$ significant improvements on all metrics compared to a single-task model using $47\%$ fewer model parameters.} 
\label{tab:table2}
\vspace{-0.3cm}
\end{table}

\begin{table}[htb]
\centering
\resizebox{.8\linewidth}{!}{%
\begin{tabular}{@{}lccccc@{}}
\toprule
 &
  \multicolumn{1}{c}{Model} &
  \begin{tabular}[c]{@{}c@{}}$\Delta_{Acc} \uparrow$\\ (\%)\end{tabular} &
  \begin{tabular}[c]{@{}c@{}}$\Delta_{M-F1} \uparrow$ \\ (\%)\end{tabular} &
  \begin{tabular}[c]{@{}c@{}}$\Delta_{W-F1} \uparrow$ \\ (\%)\end{tabular} &
  \begin{tabular}[c]{@{}c@{}}$\Delta_{\#Params} \downarrow$ \\ (\%)\end{tabular} \\ \midrule
\multirow{3}{*}{Supervised}      & DeepSense     &  -    & -    & -    & - \\
                                 & IteMM         & -45\%         & -45\%         & -44\%         & -        \\
                                 & DMTMV         & 17\%    & 17\%          & 18\%          & {\ul -20\%}   \\ \midrule
\multirow{4}{*}{Semi-Supervised} & regMVMT       & 3\%           & 4\%           & 4\%           & -0\%    \\
                                 & CSL-MTMV      & 7\%           & 7\%          & 8\%          & -0\%    \\
                                 & MFMs          & {\ul 18\%}          & {\ul 17\%}    & {\ul 19\%}    & -0\%    \\
                                 & {\it{ASM2TV}} (ours) & \textbf{35\%} & \textbf{34\%} & \textbf{33\%} & \textbf{-47\%}   \\ \bottomrule
\end{tabular}%
}
\caption{Overview of  comparison on {\bf{GLEAM}} (10 tasks 6 views). 
% We also demonstrate the average prediction results of all tasks and display the overall improvement percentage through all quantitative metrics and model space complexity. 
Generally, {\it{ASM2TV}} can achieve over $30\%$ significant improvements on all metrics compared to a single-task model using $47\%$ fewer model parameters.}  
\label{tab:table3}
\vspace{-0.3cm}
\end{table}

\begin{figure}
    \centering
    % \vspace{-.5cm}
    \includegraphics[width=\linewidth]{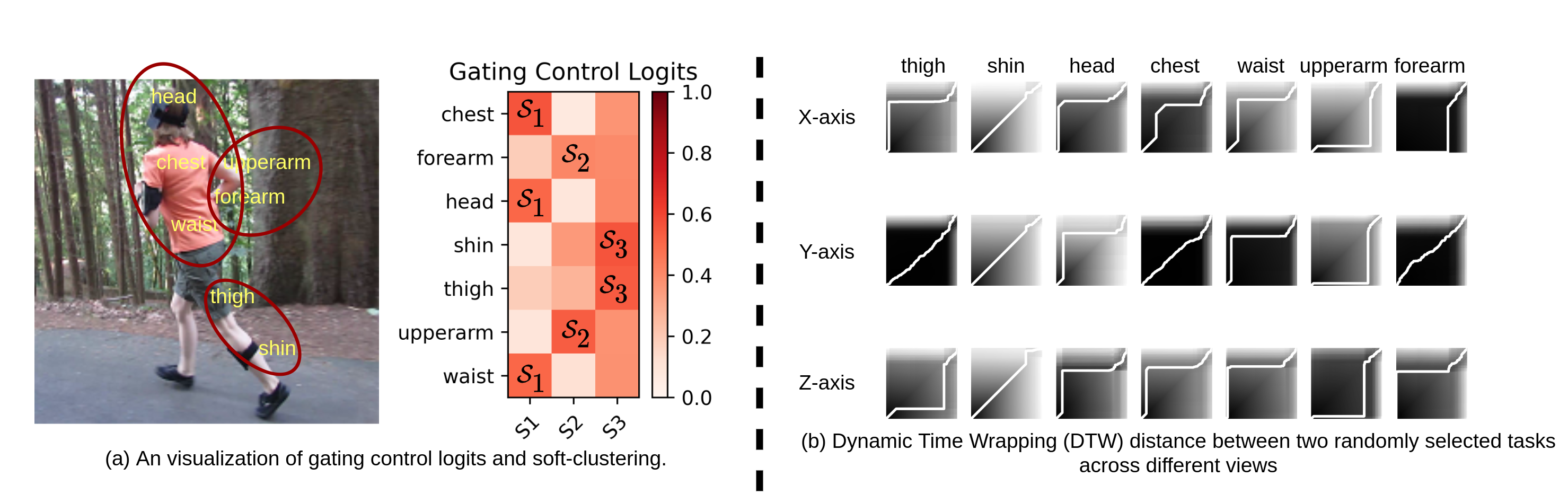}
    \caption{Visualization of correlation between different views among different tasks and the soft clustering process.}
    \label{fig:correlation}
    \vspace{-0.4cm}
\end{figure}

\subsection{Experimental Results}
\paragraph{Quantitative Results} Table \ref{tab:main_table}, \ref{tab:table4}, \ref{tab:table2} and \ref{tab:table3} show the quantitative results under three metrics for our framework and all the other competitive approaches on two datasets {\bf{RealWorld-HAR}} and {\bf{GLEAM}}. We report all metrics and relative performance of four tasks (eight tasks in total but we only exhibit four due to space limitation) in {\bf{RealWorld-HAR}} learning scenario (see Table \ref{tab:main_table}) and report a comparative performance improvements results with the single-task baseline (DeepSense) in percentage on both two datasets (see Table \ref{tab:table2} and Table \ref{tab:table3}). For more details, one may refer to the supplementary material for a full version. Specifically, we have few following observations: 1) Our ASM2TV generally achieves the best prediction performance on all metrics across all tasks under two different learning scenarios or two human action recognition datasets. 2) IteM2 is an early transductive M2TV algorithm based on graphs, originally designed for binary classification problems. It can only handle positive feature values, making it difficult to predict complex real-life mobile sensing time series data. 3) We apply the DeepSense model as the single-task learning baseline for all tasks with 10M model parameters in total and compare all the other M2TV models with it. As we discussed in Section \ref{introduction}, M2TV learning methods such as regMTMV, CSL-MTMV, and MFMs all utilize the soft-sharing mechanisms by constraining either the similarity or parameter consistency between tasks or views by keeping every model or function built on each task and each view. Thus, these models have a relatively larger parameter space since they hold the same amount of parameters as needed for single-task learning. Astonishingly, our ASM2TV inherits the spirit of hard-sharing, which achieves over $40\%$ overall improvements across all metrics while only using $47\%$ fewer parameters than the single-task learning baseline. 4) The MFMs approach generally reaches the second-best prediction results other than our framework. It is mainly due to MFMs can model a high-dimension-interacted relationship instead of either task-specific only or view-specific only relationship, using the factorization machine. However, the huge computation cost limits its power in many mobile computing scenarios. 

\paragraph{Soft-Clustering Process.}
We visualize the gating control policy logits (probabilities of open-gated for each candidate shared block) to demonstrate that this process can learn a task-view-interacted relatedness in terms of softly clustering different views from different tasks (see Figure \ref{fig:correlation}). From Figure \ref{fig:correlation}(a), different on-body positions (or views) have been automatically clustered into three groups learned by the policy. Among which, the head, chest, and waist have been clustered into $\mathcal{S}_{1}$, the upper arm and forearm have been clustered into $\mathcal{S}_{2}$ and the rest parts are assigned into the last group. To demonstrate the meaningfulness of the correlation between different body parts, we also display the Dynamic Time Wrapping (DTW) distance
% \cite{Gold2018} 
under each specific view between two randomly selected tasks. We find that this distance pattern under the head, chest, and waist is quite similar, and two observations provide a resembling conclusion. 
\begin{figure}
    \centering
    \vspace{-.5cm}
    \includegraphics[width=\linewidth]{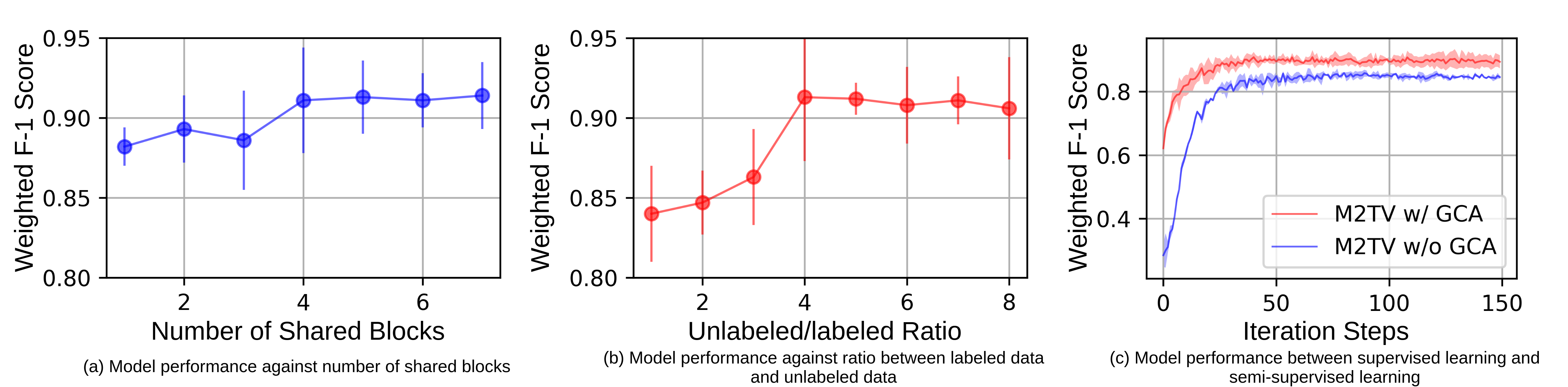}
    \caption{Ablation studies.}
    \label{fig:ablation}
    \vspace{-0.4cm}
\end{figure}

\paragraph{Ablation Studies.} We further investigate the influence of how 1) the number of shared blocks, 2) the ratio of unlabeled data to labeled data, and 3) our gathering consistency adaption can potentially affect the model performance (see Figure \ref{fig:ablation}). We have several interesting observations: 1) The overall model performance generally reaches the best when the number of shared blocks attend to 4, roughly half of the total number of views. We hypothesize that our gating control policy can find a smaller, trainable, and optimal sub-M2TV network that needs fewer shared blocks to achieve a relatively good performance rather than building a sharing scheme across tasks under every specific view. 2) For semi-supervised learning, as the proportion of unlabeled data enlarges, the model performance eventually reaches the maximum at around 4, indicating the amount of unlabeled data is four times more than labeled data. It further demonstrates our framework's effectiveness, especially under scenarios where unlabeled data is way more than labeled data such as mobile sensing in IoT. 3) Figure \ref{fig:ablation}(c) illustrates how our framework performs with or without the gathering consistency adaption procedure. It can also be viewed as supervised learning against semi-supervised learning. Our GCA procedure dramatically improves the overall prediction results by combing the unsupervised loss from fragmented time series. 

\subsubsection{Deep Insights}
 In this work, we aimed to propose a unified framework to solve the HAR problem from a different perspective: multi-task multi-view and \textit{generalize it to a wide range of mobile sensing time series applications}. Consequently, we consider HAR as a proper \textbf{representative} as we also discussed in Section \ref{introduction} that we target mobile/IoT sensing scenarios. That is also why our semi-supervised \textit{gathering consistency adaption} algorithm also targets \textbf{fragmented time series}. Though we believe our proposed \textit{sharing} scheme can be well generalized to many different scenarios, including both classification and regression datasets as MFMs \cite{Lu2017} listed, we chose to work solving HAR as the primary path of our extended research journey in multi-task multi-view learning. Our next move would be to extend our approach to evaluate more tasks such as news classification, sentiment analysis, and regression to demonstrate its effectiveness.
 
 It is imperative to handle HAR problems from the multi-task multi-view learning perspective. Because even for the same activity, the inherent patterns won't be consistent for all subjects due to the demographics. As we discussed in the Section \ref{introduction}, the heterogeneity among different views across tasks causes \textit{negative transfer} if we learn a general model using all subjects. It is also why the single model \textit{DeepSense} performs relatively worse than the other M2TV learning methods in Table \ref{tab:main_table}. Besides, the probability density function of this Gumbel-softmax concrete distribution is eventually a log-convex function of $\mathcal{X}^{t, v}$, which can be proved to be \textbf{convergent}. Once it is optimized, the sampling result for each view across each task would be \textbf{steady}, such that the choice won't be random. Additionally, the proposed gating control policy is a global learnable policy meaning it is only a $V\times N$ matrix, where $V$ is the number of views and $N$ is the number of shared blocks. The parameters of this matrix are trivial compared to the network architecture such that it won't cause extra redundancy. Once it is optimized, all the inputs from the same view of all tasks would select the same block to execute.

\section{Conclusion}
In this work, we presented a novel semi-supervised multi-task multi-view learning framework that adaptively decided which views across which tasks should share common knowledge using the gating control policy. We further showed that this gating control policy can also be viewed as a soft clustering function for different views across different tasks. Our experimental results demonstrated its  superiority by achieving the best model performance with considerably fewer model parameters than the other state-of-the-arts in the M2TVL field. Moreover, we introduced a novel semi-supervised learning method, namely gathering consistency adaption, to assist the model training on fragmented time series and enhance the overall model performance by combining the unsupervised loss. Moving forward, we would like to extend our ASM2TV to a more fine-grained architecture-wise approach 
% instead of shared-bottom only based on neural architecture search 
and explored other deep learning acceleration techniques to make our framework more efficient for mobile computing.  

\section{Acknowledgement}
This work was supported in part by the National Natural Science Foundation of China under Grant 62072279, in part by Shandong Provincial Natural Science Foundation of China under Grant ZR2021QF044.

% Use \bibliography{yourbibfile} instead or the References section will not appear in your paper
\bibliography{aaai22}

\begin{thebibliography}{39}
\providecommand{\natexlab}[1]{#1}

\bibitem[{Bachman, Alsharif, and Precup(2014)}]{Bachman2014}
Bachman, P.; Alsharif, O.; and Precup, D. 2014.
\newblock Learning with Pseudo-Ensembles.
\newblock In \emph{NeurIPS}, 3365--3373.

\bibitem[{Baxter(1997)}]{Baxter1997}
Baxter, J. 1997.
\newblock A Bayesian/Information Theoretic Model of Learning to Learn via
  Multiple Task Sampling.
\newblock \emph{Mach. Learn.}, 28(1): 7--39.

\bibitem[{Bilen and Vedaldi(2016)}]{bilen2016}
Bilen, H.; and Vedaldi, A. 2016.
\newblock Integrated perception with recurrent multi-task neural networks.
\newblock In Lee, D.~D.; Sugiyama, M.; von Luxburg, U.; Guyon, I.; and Garnett,
  R., eds., \emph{Advances in Neural Information Processing Systems 29: Annual
  Conference on Neural Information Processing Systems 2016, December 5-10,
  2016, Barcelona, Spain}, 235--243.

\bibitem[{Caruana(1998)}]{Caruana1998}
Caruana, R. 1998.
\newblock Multitask Learning.
\newblock In \emph{Learning to Learn}, 95--133.

\bibitem[{Chen et~al.(2018)Chen, Yao, Wang, Zhang, Gu, Yu, and
  Yang}]{har-iprnn}
Chen, K.; Yao, L.; Wang, X.; Zhang, D.; Gu, T.; Yu, Z.; and Yang, Z. 2018.
\newblock Interpretable parallel recurrent neural networks with convolutional
  attentions for multi-modality activity modeling.
\newblock In \emph{2018 International Joint Conference on Neural Networks
  (IJCNN)}, 1--8. IEEE.

\bibitem[{Chen et~al.(2012)Chen, Lin, Kim, Carbonell, Xing
  et~al.}]{chen2012smoothing}
Chen, X.; Lin, Q.; Kim, S.; Carbonell, J.~G.; Xing, E.~P.; et~al. 2012.
\newblock Smoothing proximal gradient method for general structured sparse
  regression.
\newblock \emph{The Annals of Applied Statistics}, 6(2): 719--752.

\bibitem[{Evgeniou and Pontil(2004)}]{evgeniou2004regularized}
Evgeniou, T.; and Pontil, M. 2004.
\newblock Regularized multi--task learning.
\newblock In \emph{SIGKDD}, 109--117.

\bibitem[{Franceschi, Dieuleveut, and Jaggi(2019)}]{Franceschi2019}
Franceschi, J.; Dieuleveut, A.; and Jaggi, M. 2019.
\newblock Unsupervised Scalable Representation Learning for Multivariate Time
  Series.
\newblock In \emph{NeurIPS}, 4652--4663.

\bibitem[{Gao et~al.(2019)Gao, Ma, Zhao, Liu, and Yuille}]{gao2019}
Gao, Y.; Ma, J.; Zhao, M.; Liu, W.; and Yuille, A.~L. 2019.
\newblock {NDDR-CNN:} Layerwise Feature Fusing in Multi-Task CNNs by Neural
  Discriminative Dimensionality Reduction.
\newblock In \emph{{IEEE} Conference on Computer Vision and Pattern
  Recognition, {CVPR} 2019, Long Beach, CA, USA, June 16-20, 2019}, 3205--3214.
  Computer Vision Foundation / {IEEE}.

\bibitem[{Guan and Pl{\"o}tz(2017)}]{mv-rnn-1}
Guan, Y.; and Pl{\"o}tz, T. 2017.
\newblock Ensembles of deep lstm learners for activity recognition using
  wearables.
\newblock \emph{Proceedings of the ACM on Interactive, Mobile, Wearable and
  Ubiquitous Technologies}, 1(2): 11.

\bibitem[{He and Lawrence(2011)}]{He2011}
He, J.; and Lawrence, R. 2011.
\newblock A Graphbased Framework for Multi-Task Multi-View Learning.
\newblock In \emph{ICML}, 25--32.

\bibitem[{Huang et~al.(2015)Huang, Feris, Chen, and Yan}]{huang2015}
Huang, J.; Feris, R.~S.; Chen, Q.; and Yan, S. 2015.
\newblock Cross-Domain Image Retrieval with a Dual Attribute-Aware Ranking
  Network.
\newblock In \emph{2015 {IEEE} International Conference on Computer Vision,
  {ICCV} 2015, Santiago, Chile, December 7-13, 2015}, 1062--1070. {IEEE}
  Computer Society.

\bibitem[{Jang, Gu, and Poole(2017)}]{Jang2017}
Jang, E.; Gu, S.; and Poole, B. 2017.
\newblock Categorical Reparameterization with Gumbel-Softmax.
\newblock In \emph{ICLR}.

\bibitem[{Jin et~al.(2013)Jin, Zhuang, Wang, He, and Shi}]{Jin2013}
Jin, X.; Zhuang, F.; Wang, S.; He, Q.; and Shi, Z. 2013.
\newblock Shared Structure Learning for Multiple Tasks with Multiple Views.
\newblock In \emph{ECML/PKDD}, volume 8189, 353--368.

\bibitem[{Jin et~al.(2014)Jin, Zhuang, Xiong, Du, Luo, and He}]{Jin2014}
Jin, X.; Zhuang, F.; Xiong, H.; Du, C.; Luo, P.; and He, Q. 2014.
\newblock Multi-task Multi-view Learning for Heterogeneous Tasks.
\newblock In \emph{CIKM}, 441--450.

\bibitem[{Kang, Grauman, and Sha(2011)}]{Kang2011}
Kang, Z.; Grauman, K.; and Sha, F. 2011.
\newblock Learning with Whom to Share in Multi-task Feature Learning.
\newblock In \emph{ICML}, 521--528.

\bibitem[{Kendall, Gal, and Cipolla(2018)}]{Kendall2018}
Kendall, A.; Gal, Y.; and Cipolla, R. 2018.
\newblock Multi-Task Learning Using Uncertainty to Weigh Losses for Scene
  Geometry and Semantics.
\newblock In \emph{CVPR}, 7482--7491.

\bibitem[{Kim and Xing(2010)}]{kim2010tree}
Kim, S.; and Xing, E.~P. 2010.
\newblock Tree-guided group lasso for multi-task regression with structured
  sparsity.
\newblock In \emph{ICML}, volume~2, 1.

\bibitem[{Kokkinos(2017)}]{kokkinos2017}
Kokkinos, I. 2017.
\newblock UberNet: Training a Universal Convolutional Neural Network for Low-,
  Mid-, and High-Level Vision Using Diverse Datasets and Limited Memory.
\newblock In \emph{2017 {IEEE} Conference on Computer Vision and Pattern
  Recognition, {CVPR} 2017, Honolulu, HI, USA, July 21-26, 2017}, 5454--5463.
  {IEEE} Computer Society.

\bibitem[{Laine and Aila(2017)}]{Laine2017}
Laine, S.; and Aila, T. 2017.
\newblock Temporal Ensembling for Semi-Supervised Learning.
\newblock In \emph{ICLR}.

\bibitem[{Lu et~al.(2017)Lu, He, Shao, Cao, and Yu}]{Lu2017}
Lu, C.; He, L.; Shao, W.; Cao, B.; and Yu, P.~S. 2017.
\newblock Multilinear Factorization Machines for Multi-Task Multi-View
  Learning.
\newblock In \emph{WSDM}, 701--709.

\bibitem[{Ma et~al.(2019)Ma, Li, Zhang, Gao, and Lu}]{attnsense}
Ma, H.; Li, W.; Zhang, X.; Gao, S.; and Lu, S. 2019.
\newblock AttnSense: multi-level attention mechanism for multimodal human
  activity recognition.
\newblock In \emph{Proceedings of the 28th International Joint Conference on
  Artificial Intelligence}, 3109--3115. AAAI Press.

\bibitem[{Maddison, Tarlow, and Minka(2014)}]{Maddison2014}
Maddison, C.~J.; Tarlow, D.; and Minka, T. 2014.
\newblock A* Sampling.
\newblock In \emph{NeurIPS}, 3086--3094.

\bibitem[{Mahdavinejad et~al.(2018)Mahdavinejad, Rezvan, Barekatain, Adibi,
  Barnaghi, and Sheth}]{Mahdavinejad2018}
Mahdavinejad, M.~S.; Rezvan, M.; Barekatain, M.; Adibi, P.; Barnaghi, P.~M.;
  and Sheth, A.~P. 2018.
\newblock Machine learning for Internet of Things data analysis: {A} survey.
\newblock \emph{CoRR}, abs/1802.06305.

\bibitem[{Misra et~al.(2016)Misra, Shrivastava, Gupta, and Hebert}]{Misra2016}
Misra, I.; Shrivastava, A.; Gupta, A.; and Hebert, M. 2016.
\newblock Cross-Stitch Networks for Multi-task Learning.
\newblock In \emph{CVPR}, 3994--4003.

\bibitem[{Rahman et~al.(2015)Rahman, Merck, Huang, and Kleinberg}]{Rahman2015}
Rahman, S.~A.; Merck, C.~A.; Huang, Y.; and Kleinberg, S. 2015.
\newblock Unintrusive eating recognition using Google Glass.
\newblock In \emph{9th ICPCTH}, 108--111.

\bibitem[{Ranjan, Patel, and Chellappa(2019)}]{ranjan2019}
Ranjan, R.; Patel, V.~M.; and Chellappa, R. 2019.
\newblock HyperFace: {A} Deep Multi-Task Learning Framework for Face Detection,
  Landmark Localization, Pose Estimation, and Gender Recognition.
\newblock \emph{{IEEE} Trans. Pattern Anal. Mach. Intell.}, 41(1): 121--135.

\bibitem[{Rasmus et~al.(2015)Rasmus, Berglund, Honkala, Valpola, and
  Raiko}]{Rasmus2015}
Rasmus, A.; Berglund, M.; Honkala, M.; Valpola, H.; and Raiko, T. 2015.
\newblock Semi-supervised Learning with Ladder Networks.
\newblock In \emph{NeurIPS}, 3546--3554.

\bibitem[{Ruder(2017)}]{Ruder2017}
Ruder, S. 2017.
\newblock An Overview of Multi-Task Learning in Deep Neural Networks.
\newblock \emph{CoRR}, abs/1706.05098.

\bibitem[{Ruder et~al.(2019)Ruder, Bingel, Augenstein, and
  S{\o}gaard}]{ruder2019}
Ruder, S.; Bingel, J.; Augenstein, I.; and S{\o}gaard, A. 2019.
\newblock Latent Multi-Task Architecture Learning.
\newblock In \emph{The Thirty-Third {AAAI} Conference on Artificial
  Intelligence, {AAAI} 2019, The Thirty-First Innovative Applications of
  Artificial Intelligence Conference, {IAAI} 2019, The Ninth {AAAI} Symposium
  on Educational Advances in Artificial Intelligence, {EAAI} 2019, Honolulu,
  Hawaii, USA, January 27 - February 1, 2019}, 4822--4829. {AAAI} Press.

\bibitem[{Standley et~al.(2020)Standley, Zamir, Chen, Guibas, Malik, and
  Savarese}]{Standley2020}
Standley, T.; Zamir, A.~R.; Chen, D.; Guibas, L.~J.; Malik, J.; and Savarese,
  S. 2020.
\newblock Which Tasks Should Be Learned Together in Multi-task Learning?
\newblock In \emph{ICML}, volume 119, 9120--9132.

\bibitem[{Sun et~al.(2020)Sun, Panda, Feris, and Saenko}]{Sun2020}
Sun, X.; Panda, R.; Feris, R.; and Saenko, K. 2020.
\newblock AdaShare: Learning What To Share For Efficient Deep Multi-Task
  Learning.
\newblock In \emph{NeurIPS}.

\bibitem[{Sztyler and Stuckenschmidt(2016)}]{Sztyler2016}
Sztyler, T.; and Stuckenschmidt, H. 2016.
\newblock On-body localization of wearable devices: An investigation of
  position-aware activity recognition.
\newblock In \emph{2016 {IEEE} International Conference on Pervasive Computing
  and Communications}, 1--9.

\bibitem[{Wu, Zhan, and Jiang(2018)}]{Wu2018}
Wu, Y.; Zhan, D.; and Jiang, Y. 2018.
\newblock {DMTMV:} {A} Unified Learning Framework for Deep Multi-task
  Multi-view Learning.
\newblock In \emph{ICBK}, 49--56.

\bibitem[{Xie et~al.(2020)Xie, Dai, Hovy, Luong, and Le}]{Xie2020}
Xie, Q.; Dai, Z.; Hovy, E.~H.; Luong, T.; and Le, Q. 2020.
\newblock Unsupervised Data Augmentation for Consistency Training.
\newblock In \emph{NeurIPS}.

\bibitem[{Yao et~al.(2017{\natexlab{a}})Yao, Hu, Zhao, Zhang, and
  Abdelzaher}]{deepSense}
Yao, S.; Hu, S.; Zhao, Y.; Zhang, A.; and Abdelzaher, T. 2017{\natexlab{a}}.
\newblock Deepsense: A unified deep learning framework for time-series mobile
  sensing data processing.
\newblock In \emph{Proceedings of the 26th International Conference on World
  Wide Web}, 351--360. International World Wide Web Conferences Steering
  Committee.

\bibitem[{Yao et~al.(2017{\natexlab{b}})Yao, Hu, Zhao, Zhang, and
  Abdelzaher}]{Yao2017}
Yao, S.; Hu, S.; Zhao, Y.; Zhang, A.; and Abdelzaher, T.~F. 2017{\natexlab{b}}.
\newblock DeepSense: {A} Unified Deep Learning Framework for Time-Series Mobile
  Sensing Data Processing.
\newblock In \emph{WWW}, 351--360.

\bibitem[{Zhang and Huan(2012)}]{Zhang2012}
Zhang, J.; and Huan, J. 2012.
\newblock Inductive multi-task learning with multiple view data.
\newblock In \emph{SIGKDD}, 543--551.

\bibitem[{Zhang et~al.(2018)Zhang, Li, Nguyen, Zhuang, Xiong, and
  Lu}]{zhang2018label}
Zhang, X.; Li, W.; Nguyen, V.; Zhuang, F.; Xiong, H.; and Lu, S. 2018.
\newblock Label-Sensitive Task Grouping by Bayesian Nonparametric Approach for
  Multi-Task Multi-Label Learning.
\newblock In \emph{IJCAI}, 3125--3131.

\end{thebibliography}

% \section{Acknowledgments}
% AAAI is especially grateful to Peter Patel Schneider for his work in implementing the original aaai.sty file, liberally using the ideas of other style hackers, including Barbara Beeton. We also acknowledge with thanks the work of George Ferguson for his guide to using the style and BibTeX files --- which has been incorporated into this document --- and Hans Guesgen, who provided several timely modifications, as well as the many others who have, from time to time, sent in suggestions on improvements to the AAAI style. We are especially grateful to Francisco Cruz, Marc Pujol-Gonzalez, and Mico Loretan for the improvements to the Bib\TeX{} and \LaTeX{} files made in 2020.

% The preparation of the \LaTeX{} and Bib\TeX{} files that implement these instructions was supported by Schlumberger Palo Alto Research, AT\&T Bell Laboratories, Morgan Kaufmann Publishers, The Live Oak Press, LLC, and AAAI Press. Bibliography style changes were added by Sunil Issar. \verb+\+pubnote was added by J. Scott Penberthy. George Ferguson added support for printing the AAAI copyright slug. Additional changes to aaai22.sty and aaai22.bst have been made by Francisco Cruz, Marc Pujol-Gonzalez, and Mico Loretan.

% \bigskip
% \noindent Thank you for reading these instructions carefully. We look forward to receiving your electronic files!
\appendix

\section{Appendix}
\subsection{Datasets and Descriptions}
\paragraph{RealWorld-HAR.}The data set\footnote{https://sensor.informatik.uni-mannheim.de/\#dataset\_realworld} covers acceleration, GPS, gyroscope, light, magnetic field, and sound level data of the activities climbing stairs down and up, jumping, lying, standing, sitting, running/jogging, and walking of fifteen subjects (age 31.9±12.4, height 173.1±6.9, weight 74.1±13.8, eight males and seven females). \textbf{We select 4 males and 4 females out of the 15 subjects for experiments.} For each activity, we recorded simultaneously the acceleration of the body positions chest, forearm, head, shin, thigh, upper arm, and waist. Each subject performed each activity roughly 10 minutes except for jumping due to the physical exertion (~1.7 minutes). Concerning male and female, the amount of data is equally distributed. Each movement was recorded by a video camera to facilitate the usage. \textbf{Since we treat the seven different on-body positions as seven views, we then have 8 tasks with 7 views defined on this dataset.}
\paragraph{GLEAM.} Basic features 2 hours of high resolution activity data for each of 38 participants as they walk, talk, and eat meals, collected with Google Glass\footnote{http://www.healthailab.org/data.html}. The labeled data from this head-mounted sensor can be used to recognize eating and other activities, and ultimately to assist individuals with chronic disease (e.g. insulin dosing and glucose testing reminders for people with diabetes). Head movement has been underappreciated as a data source for eating and activity recognition, so this study aims to address this by creating a public data resource that contains labeled data from human head movement. Data was collected from all of Glass’s sensors (accelerometer, gravity, linear
acceleration, rotation, gyroscope, magnetic field, and light). The table on the following page gives brief descriptions of the sensors. We aimed to record data at a resolution of 200 milliseconds. However, due to some system processes and their higher priorities, data recording was occasionally preempted so the median interval between recordings
was 395 milliseconds. \textbf{We treat each sensor as each single view, thus we have a total of 38 tasks with 6 views each defined on this dataset.}

\subsection{Experimental Settings}
For IteM2, we set $\alpha$, a taskNum-by-viewNum matrix, $\alpha_{task,view}$ is set to 1 in the original paper, $\mu$ a vector of $\mu$ for each task, in which $\mu_{task}=0.01$ in the as set originally; the number of iteration steps is set to 100; for CSL-MTMV, $b$ is set to 1, the number of iterations is 10, the dimension of shared low dimensional feature space is set to 20, task relation regularization parameter $\alpha$ is set to 0.01, $\beta$ is set to 0.01, $\gamma$ is also set to 0.01; for regMTMV, it has the same parameter settings with CSL-MTMV; for MFMs, $\lambda$ and $\gamma$ are the regularization parameters, and $\lambda$ is set to $1e^{-3}$, $\gamma$ is set to $1e^{-3}$, the dimension of latent factors $R=20$, learning rate is initialized by 0.1, $\sigma=1$, the maximum number of iterations are all set to 200. For DMTMV, we set batch size to 512, learning rate to 0.1, and a regularization parameter $\lambda$ to $1e^{-6}$. For our $ASM2TV$, we set the batch size of labeled train data as 16, 24 for unlabeld data and 32 for evaluation and test data. For any deep neural network based approaches, we use Pytorch\footnote{https://pytorch.org} as our coding framework. We conduct all experiments on a single NVIDIA Tesla P100 GPU. 

\end{document}